\documentclass[conference]{IEEEtran}
\pdfoutput=1
\usepackage{ifpdf}
\usepackage{cite}
\usepackage{amsmath,amssymb,amsfonts}
\usepackage{algorithm}
\usepackage{subfigure}
\usepackage{algpseudocode}
\usepackage{graphicx}
\usepackage{textcomp}
\usepackage{xcolor}
\usepackage{tabularx}
\usepackage{verbatim}
\usepackage{multirow}
\usepackage{balance}
\usepackage{hyperref}
\usepackage{epstopdf}
\usepackage[T1]{fontenc}
\usepackage[utf8x]{inputenc}

\def\BibTeX{{\rm B\kern-.05em{\sc i\kern-.025em b}\kern-.08em
    T\kern-.1667em\lower.7ex\hbox{E}\kern-.125emX}}
\begin{document}

\title{Using 3D Shadows to Detect Object Hiding Attacks on Autonomous Vehicle Perception \\

}


\author{\IEEEauthorblockN{Zhongyuan Hau}
\IEEEauthorblockA{\textit{Department of Computing} \\
\textit{Imperial College London}\\
zy.hau17@imperial.ac.uk}
\and

\IEEEauthorblockN{Soteris Demetriou}
\IEEEauthorblockA{\textit{Department of Computing} \\
\textit{Imperial College London}\\
s.demetriou@imperial.ac.uk}
\and

\IEEEauthorblockN{Emil C. Lupu
}
\IEEEauthorblockA{\textit{Department of Computing} \\
\textit{Imperial College London}\\
e.c.lupu@imperial.ac.uk}

}

\maketitle
\begin{abstract}
Autonomous Vehicles (AVs) are mostly reliant on LiDAR sensors which enable spatial perception of their surroundings and help make driving decisions. Recent works demonstrated attacks that aim to hide objects from AV perception, which can result in severe consequences. 3D shadows, are regions void of measurements in 3D point clouds which arise from occlusions of objects in a scene. 3D shadows were proposed as a physical invariant valuable for detecting spoofed or fake objects. In this work, we leverage 3D shadows to locate obstacles that are hidden from object detectors. We achieve this by searching for void regions and locating the obstacles that cause these shadows. Our proposed methodology can be used to detect an object that has been hidden by an adversary as these objects, while hidden from 3D object detectors, still induce shadow artifacts in 3D point clouds, which we use for obstacle detection. We show that using 3D shadows for obstacle detection can achieve high accuracy in matching shadows to their object and provide precise prediction of an obstacle's distance from the ego-vehicle.
\end{abstract}


\section{Introduction}
Autonomous vehicles (AVs) are increasingly being deployed on public roads. AV's perception of the environment they operate in is fundamental to their autonomy, guiding their driving decisions for safe and reliable operation. These vehicles are commonly equipped with LiDAR sensors, which collect high definition depth measurements stored in 3D point clouds. Objects detected from the 3D point clouds are used by AVs to map the obstacles in the driving environment, which is vital to the safety of the autonomous vehicle, its passengers and surroundings. Object hiding attacks have been explored in various works \cite{tu2020physically, cao2019adversarialobjects, cao2021invisible, hau2021object} which demonstrate that it is possible to generate adversarial objects that can evade object detection. Object detection is a safety-critical function and failing to detect an object can lead to fatal collisions.

\vspace{3pt}\noindent\textbf{Object hiding attacks.} There are currently 2 classes of Object Hiding Attacks. The first type is when the adversarial operations are performed on the physical (target) object itself. One proposed method \cite{tu2020physically} is to place adversarial objects on top of a target vehicle that evade point-cloud based object detectors, achieving attack success rate of 80\%. The adversarial objects can be generated in a white-box or a black-box method. For the white-box attack, the adversarial object is generated using a gradient-based approach to minimize the confidence score of the target object (vehicle); whereas the black-box attack chooses adversarial objects using a genetic algorithm approach to iterate and improve on adversarial object meshes. Cao et al. in \cite{cao2019adversarialobjects} proposed using an optimization-based approach to generate adversarial objects that evade detection by 3D object detectors. In \cite{cao2021invisible}, the authors exploited the weakness of deep neural-networks (DNN) based detectors and proposed an optimization-based method to generate physically realizable adversarial objects that target multi-sensor fusion, effectively evading both Camera and LiDAR based detectors.

The second type of object hiding attacks is when adversarial operations are performed on the sensing modality itself. In \cite{hau2021object}, the authors proposed using the LiDAR spoofing approach (Object Removal Attack, \textit{ORA-Random}) to inject random points in the target's object bounding box to displace points in the original object's point cloud, causing mis-detection of the target object. They showed that ORA was effective in damaging the performance of object detectors for front-near objects and works well for both large objects such as cars and small objects such as pedestrians and cyclists.

\vspace{3pt}\vspace{3pt}\noindent\textbf{Motivation.} Whilst many works demonstrate different ways to hide objects from detection by object detectors, there is a lack of research into how we can reliably detect such object hiding attacks. Cao et al. in \cite{cao2021invisible} showed that model-level defenses (i.e. input transformations and adversarial training) on multi-sensor fusion models to increase their adversarial robustness does not effectively decrease the attack success rates. The authors have also suggested fusing more sensor modalities to increase the robustness, but argue that this approach does not fundamentally address the weakness of using DNN-based object detectors and would still be exploitable. To this end, we develop an object hiding attack detection methodology that leverages a strong physical invariant of a single sensing modality (LiDAR) to provide an orthogonal and reliable detection of objects in a 3D scene, which is robust against object hiding attacks.

\begin{figure*}[htb]
    \centering
    \includegraphics[width=0.85\textwidth, clip]{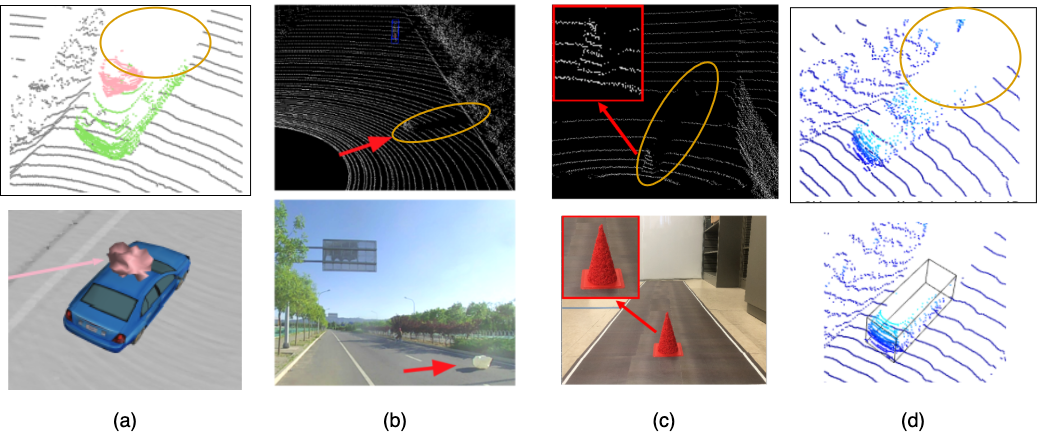}
    \caption{Object Hiding Attacks from: (a) Tu et al.\cite{tu2020physically} (b) Cao et al.\cite{cao2019adversarial} (c) Cao et al.\cite{cao2021invisible} (d) Hau et al.\cite{hau2021object}. Shadows of adversarial objects circled in orange.}
    \label{fig:shadow_oha}
    \vspace{-3mm}
\end{figure*}

\vspace{3pt}\vspace{3pt}\noindent\textbf{Our work.} 3D shadows in LiDAR point clouds, introduced in \cite{Hau2021Shadow}, are a physical phenomenon that is caused by occlusion of LiDAR laser pulses by objects in a scene. The authors leverage these shadow artifacts as a physical invariant to detect LiDAR spoofing attacks. We observed in Figure \ref{fig:shadow_oha}, that under object hiding attacks, the objects hidden by the adversary in the scene, are not detected by a DNN-based object detector, but still exist and occlude LiDAR pulses to leave behind shadow artifacts in the 3D point cloud. We leverage this observation and propose an obstacle detection methodology that searches a region of interest in a 3D point cloud for objects based on the shadow artifacts found in the 3D point cloud scene. Our proposed methodology of using shadows provides an alternate approach to detecting objects in a 3D scene and can be used as an orthogonal defense against object hiding attacks.

\section{Threat Model}
\label{sec:threat_model}
In this work, we investigate the detection of unidentified objects using their shadow artifacts in a 3D point cloud. Locating unidentified objects can help us detect Object Hiding Attacks, a class of attacks on AV perception that aims to cause mis-detection of objects in the sensed environment.

\vspace{3pt}\noindent\textbf{Threat model.} We assume an adversary $\mathcal{A}$, that has the ability to perform any of the state-of-the-art object hiding attacks. For Class 1 Object Hiding Attacks, there are no perturbations to the 3D LiDAR point cloud, and thus, the observation that object shadows exist holds. For Class 2, Object Removal Attack (\textit{ORA-Random}) \cite{hau2021object} generate adversarial objects by perturbing the point cloud. However, the point perturbations are limited to a short distance of point shifting along the ray direction and do not affect the object's shadow. 

\vspace{3pt}\noindent\textbf{Emulating object hiding attacks.} Object Hiding Attacks are thus observed to not alter the 3D shadows of the adversarial objects. Additionally, we also assume an adversary that does not perturb the shadows. As such, to emulate such object hiding attack for the evaluation of our proposed detection mechanism, we simply ignore that these objects are detected by an object detector and remove the labelled object from any output. In short, we assume that the attacker has successfully conducted an object hiding attack and the output labels do not contain the target object.

\section{Using 3D Shadows to Detect Hidden Objects}
\label{sec:obj_shadow}
In this work, we use the key observation that 3D shadows are caused by occlusion of LiDAR laser pulses by opaque objects. Adversarial objects in a 3D point cloud scene that are subjected to Object Hiding Attacks, would not be detected by an DNN-based object detector, but due to the physics of LiDAR, the phenomenon of 3D shadows would still exist. The existence of the objects' 3D shadows in a 3D point cloud can be exploited to locate and identify adversarial objects that evade object detection.

\vspace{3pt}\noindent\textbf{3D Shadows as physical invariant.} 3D shadows have been introduced by Hau et al. in \cite{Hau2021Shadow} as a phyiscal invariant to verify genuine objects in a 3D point cloud scene and detect object spoofing attacks. In object spoofing attacks \cite{petit2015remote, shin2017illusion, cao2019adversarial, 255240}, point cloud measurements are injected into a scene to spoof objects that are then erroneously detected by 3D-object detectors used by AVs. Shadow-Catcher \cite{hau2021object} detects such attacks by using the bounding box of objects generated by object detectors to identify a shadow region behind the detected object and inspects that shadow region for presence of point cloud measurements, which are indicative of a spoofing attack. The authors have shown that 3D shadows are a strong physical invariant that can be used to verify genuine objects. In this work, we explore searching the scene for void regions and attributing them to genuine objects.

\vspace{3pt}\noindent\textbf{3D Shadows to detect object hiding attacks.} From this observation that 3D shadows are strong a physical invariant that can be used to verify genuine objects in a 3D scene, we propose a methodology that identifies void regions in a 3D scene, attributes it to objects and corroborates this with the output of object detectors to: 1) verify genuine objects and 2) detect potential object hiding attacks in a scene.

\section{Obstacle Detection Design}
\label{sec:system}
We propose a methodology that searches a region of interest in the 3D point cloud scene for void regions, and analyzes them to either verify that the void region is caused by the occlusion of a detected object or an unidentified obstacle that could be an adversarial hidden object. This approach is different from Shadow-Catcher \cite{Hau2021Shadow} as Shadow-Catcher uses the bounding box of a detected object to obtain the shadow region for analysis to detect spoofed objects. With the Object Hiding Attack adversary we consider that the attacker aims to hide the object and thus there is no bounding box for Shadow-Catcher to perform analysis on -- Shadow-Catcher is blind to an object hiding adversary. As our approach is unable to identify the objects (i.e. classify and label the object), in the rest of the paper, we call the objects detected by our approach as \textit{\textbf{obstacles}}. Objects hidden by an adversary will still have shadows in a 3D point cloud, and thus can still be detected as an obstacle by our methodology.

\vspace{3pt}\vspace{3pt}\noindent\textbf{Identifying shadow regions.} We first define a region of interest (RoI) to analyse for shadows. We look into the front-near region of the ego-vehicle and define a region of width 10m and length 30m in front of it, where we analyse for shadows and subsequently identify obstacles. Next, from the RoI, a thin volume of the point cloud around the ground level is extracted for analysis. The RoI volume is discretized into cubic cells with width of 0.3m, and the occupancy of these cells by LiDAR point measurements is checked. Cells that are empty are clustered using DBSCAN \cite{ester1996density} into shadow clusters.

The parameters used such as RoI size and cubic cell width can be tuned based on the visibility requirements (i.e. a larger RoI for greater visibility) and the resolution of the LiDAR (a LiDAR that uses less lasers would have a lower resolution and thus, requires a larger cell width for analysis).

\vspace{3pt}\vspace{3pt}\noindent\textbf{Attributing shadows to their obstacles and detected object.} In a simple case where objects in a scene are sparsely located, each shadow cluster would correspond to a single obstacle. However, in crowded scenes, the shadows of the objects overlap resulting in a single large cluster. In order to distinguish the different obstacles that contribute to the shadow, we use the center of the cells in each shadow cluster to generate a collection of frustums to the LiDAR on the ego-vehicle (origin coordinates of the point cloud). The points in the collection of frustums are then checked to determine whether they fall within bounding boxes of objects identified by 3D Object Detectors. The remaining points that do not fall in the bounding box of detected objects are then further processed to identify and localize the obstacles.

\vspace{3pt}\vspace{3pt}\noindent\textbf{Localizing unidentified obstacles.} The remaining points (that do not belong to any identified objects) from the frustums of shadows are further processed as follows. The points undergo clustering with DBSCAN where we get clusters of points that according to the physics of LiDAR, contribute to the formation of shadows and are indicative of an opaque obstacle in the scene. Finally, we generate individual bounding boxes for each cluster that enclose their points.

\subsection{Example 1: Single Object in Front-Near Region}\label{subsec:ex1}
\begin{figure}[htb]
    \centering
    \subfigure[Image of Scene with 1 Object (Car) In Front-Near Region]
    {
        \includegraphics[width=0.45\textwidth]{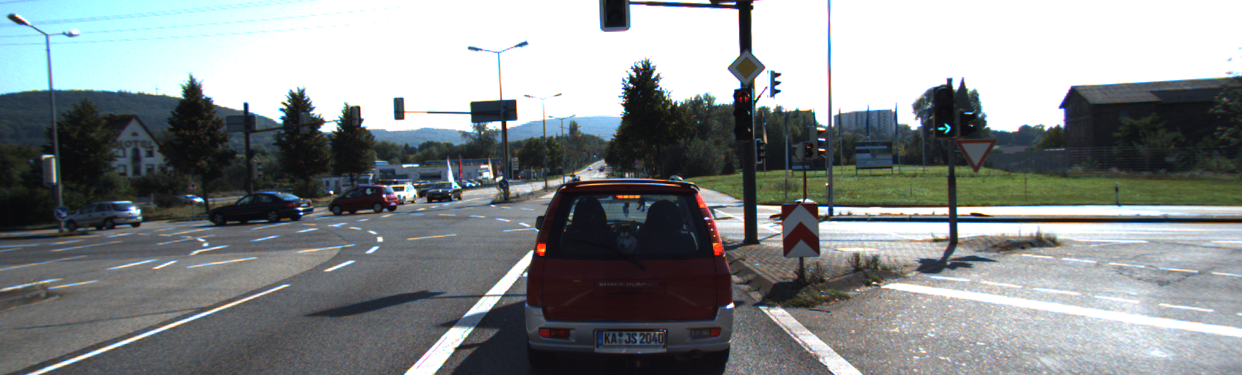}
        \label{fig:scene_1}
    }
    \subfigure[BEV of 3D point cloud with 1 object in Front-Near Region]
    {
        \includegraphics[width=0.45\textwidth]{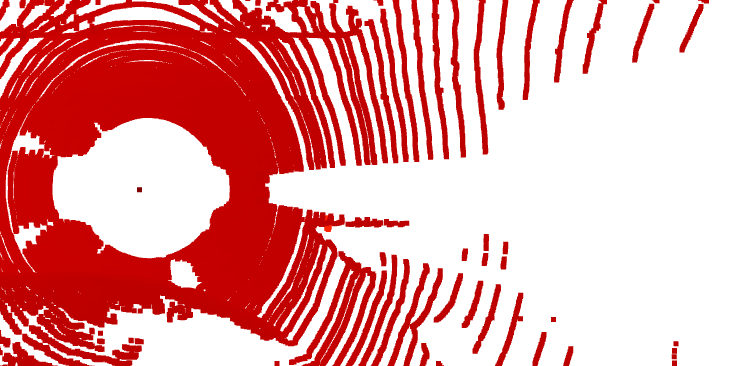}
        \label{fig:scene_1_bev}
    }
    \caption{Scene with 1 Object (Car) In Front-Near Region}
\end{figure}

We first demonstrate the use of the proposed obstacle detection in a simple case where the scene (Fig. \ref{fig:scene_1}) has a single Car object in the front-near region of ego-vehicle, and the birds-eye-view (BEV) of the 3D point cloud in Fig. \ref{fig:scene_1_bev}. 

\begin{figure}[htb]
    \centering
    \includegraphics[width=0.45\textwidth]{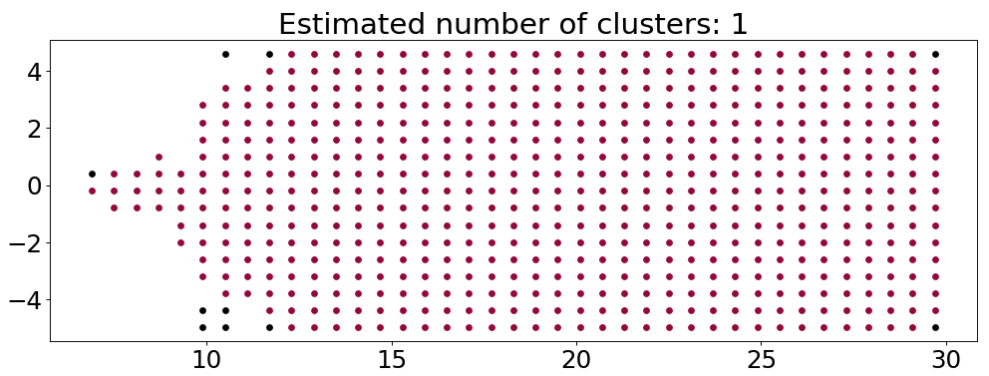}
    \caption{Clustering of empty cells into shadow clusters -- 1 Shadow Cluster. The origin (0,0) is the LiDAR unit on the ego-vehicle.}
    \label{fig:scene_1_shadow_cluster}
\end{figure}

\begin{figure}[htb]
    \centering
    \subfigure[Frustums from Shadow Clusters to LiDAR]
    {
        \includegraphics[width=0.45\textwidth]{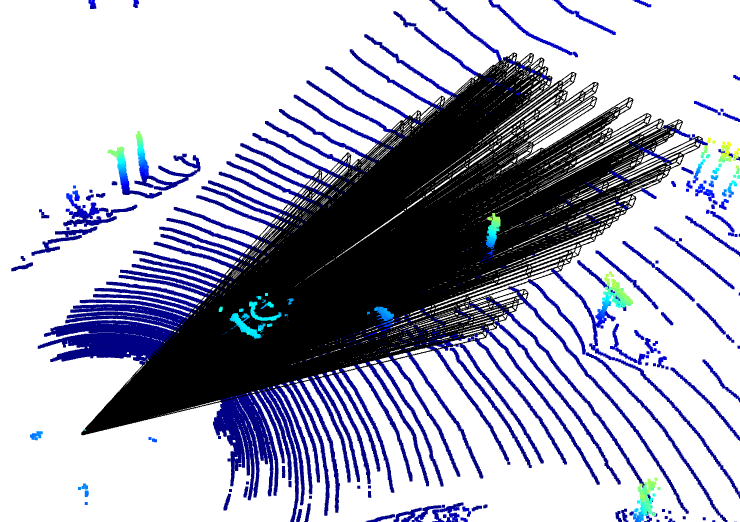}
        \label{fig:scene_1_frustums}
    }
    \subfigure[Clusters of points in frustums]
    {
        \includegraphics[width=0.45\textwidth]{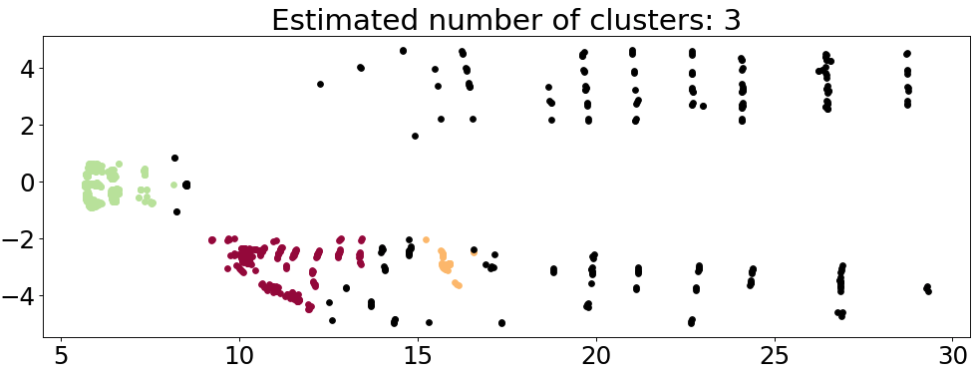}
        \label{fig:scene_1_obstacle_clusters}
    }
    \subfigure[Bounding Boxes of Obstacles identified from clusters]
    {
        \includegraphics[width=0.45\textwidth]{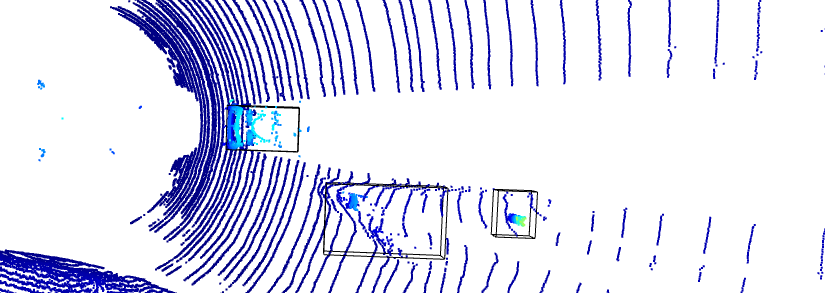}
        \label{fig:scene_1_obstacle_bbox}
    }
    \caption{Processing of Shadow Clusters into Obstacles}
\end{figure}

We use our proposed Obstacle Detection methodology and first search for void regions in the scene by discretizing the RoI into cells and checking if the cells are occupied by points. The empty cells are clustered and the result is shown in Fig. \ref{fig:scene_1_shadow_cluster}. Frustums are then generated (Fig. \ref{fig:scene_1_frustums} from the LiDAR to the cells in the shadow clusters, where the points in the frustums are clustered as obstacles (Fig. \ref{fig:scene_1_obstacle_clusters}). Finally, the clusters of points, identified as obstacles in the scene, are enclosed with bounding boxes (Fig. \ref{fig:scene_1_obstacle_bbox}).

Our proposed detection methodology was able to accurately detect the car in-front of the ego-vehicle as an obstacle and in addition, picked up two other obstacles which are the road divider and the traffic light post.

\subsection{Example 2: Multiple Objects in Front-Near Region}
\begin{figure}[htb]
    \centering
    \includegraphics[width=0.45\textwidth, clip]{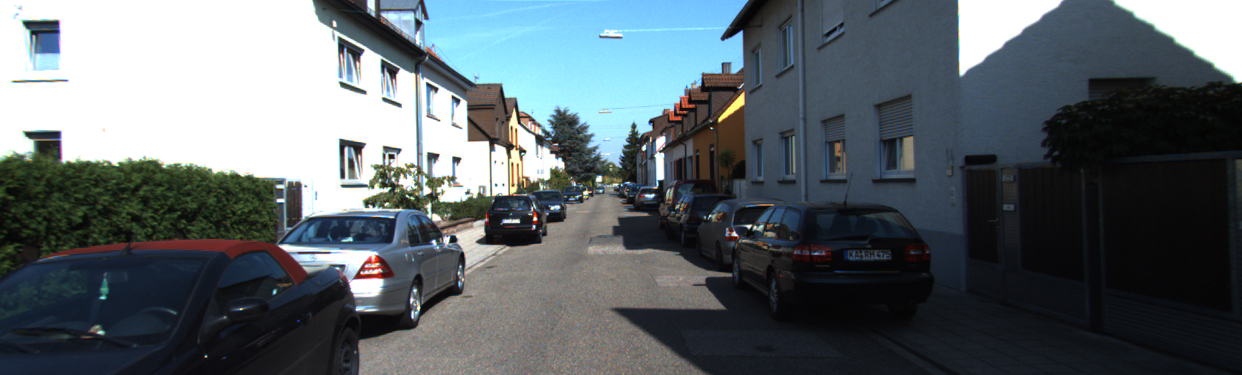}
    \caption{Image of Scene with Multiple Objects In Front-Near Region.}
    \label{fig:scene_2}
    \vspace{-5mm}
\end{figure}

Next, we demonstrate the effectiveness of the proposed obstacle detection methodology in a scene where there are multiple objects in the front-near region of the ego-vehicle. The scene is shown in Fig. \ref{fig:scene_2}, where there are multiple cars parked along the side of the road.

\begin{figure}[htb]
    \centering
    \vspace{-3.5mm}
    \subfigure[Frustums from Shadow Clusters to LiDAR]
    {
        \includegraphics[width=0.45\textwidth]{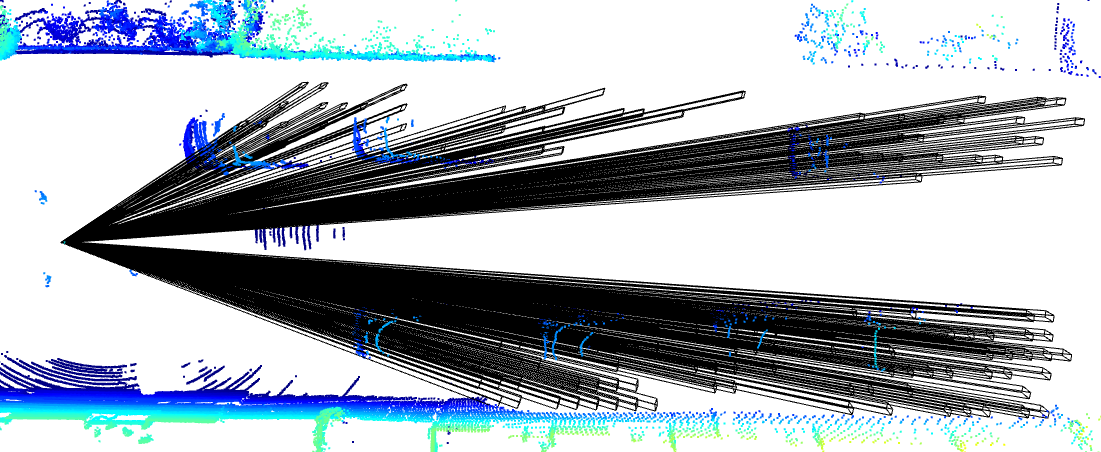}
        \label{fig:scene_2_frustums}
    }
    \subfigure[Clusters of points in frustums]
    {
        \includegraphics[width=0.45\textwidth]{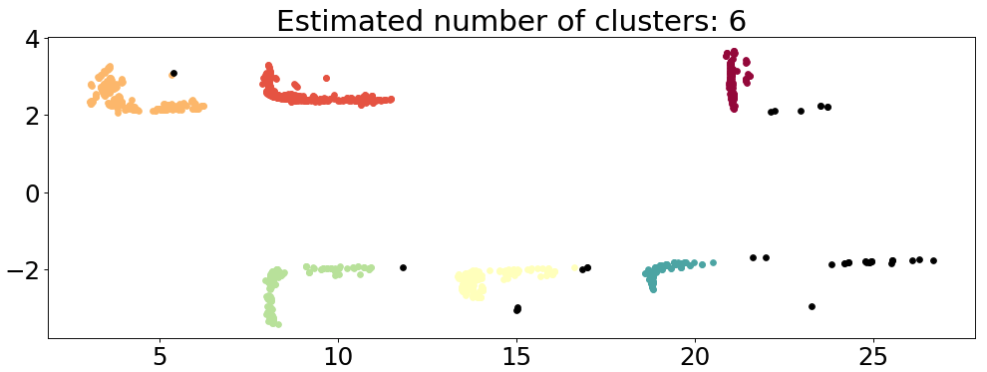}
        \label{fig:scene_2_obstacle_clusters}
    }
    \subfigure[Bounding Boxes of Obstacles identified from clusters]
    {
        \includegraphics[width=0.45\textwidth]{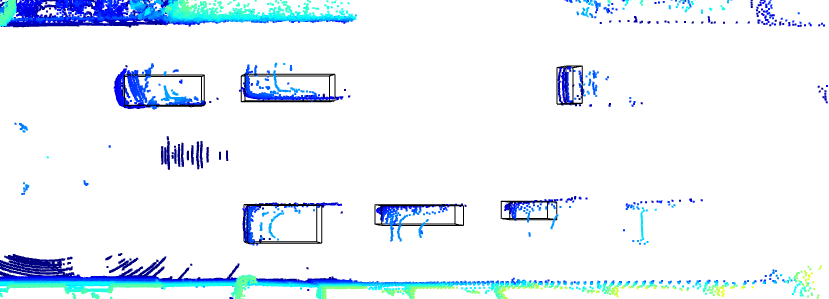}
        \label{fig:scene_2_obstacle_bbox}
    }
    \caption{Processing of Shadow Clusters into Obstacles}
    \vspace{-5mm}
\end{figure}

Using the Obstacle Detection methodology, the scene is analysed for void regions in the front-near area of interest and the void cells are clustered together to form shadow clusters. Frustums from the LiDAR to the shadow clusters are then generated (Fig. \ref{fig:scene_2_frustums}, the points in the frustums are clustered to obtain point clusters of obstacles (Fig. \ref{fig:scene_2_obstacle_clusters} and their bounding boxes are also generated (Fig. \ref{fig:scene_2_obstacle_bbox}).

From Fig. \ref{fig:scene_2_obstacle_bbox}, the obstacles identified using the proposed Obstacle Detection Methodology match well with the objects that are found in the scene (i.e. the cars that are parked along the sides of the road).

\subsection{Summary}
We have proposed an Obstacle Detection Methodology that first searches an area of interest in the front-near region of the ego-vehicle for void spaces that are indicative of shadow regions. Ray frustums from the LiDAR to the shadow clusters are generated and points that falls within these frustums are further clustered to obtain point clusters of obstacles that are found in the scene. From the demonstration of the 2 examples, we show that our proposed methodology is capable of accurately localizing obstacles in a scene. Next, we perform experiments to quantitatively measure the accuracy and effectiveness of the proposed methodology in detecting obstacles.

\section{Experiments \& Evaluation}
\label{sec:evaluation}

\vspace{3pt}\vspace{3pt}\noindent\textbf{Models \& Datasets.} We perform our evaluation of the effectiveness and efficiency of our shadow detection methodology and object verification on the KITTI dataset \cite{Geiger2013IJRR}, which contains LiDAR measurements from scenes captured in the real-world. Object Hiding Attacks are emulated by removing the target object's label from KITTI ground truth labels.

\vspace{3pt}\vspace{3pt}\noindent\textbf{Evaluation scenarios.} In our evaluation of the effectiveness and utility of the proposed object hiding attack detection methodology, we design experiments that aim to answer the following research questions:

\vspace{3pt}\vspace{3pt}\noindent\textit{RQ1: How well does the shadow identification and verification work in a benign scenario?} In this experiment, we use benign scenes from KITTI and evaluate how well the shadow identification and verification procedure by measuring the accuracy of matching shadows to ground truth labelled objects. For each identified shadow, we sample the discretized space of the shadow and obtain the frustum from the LiDAR to the shadow cell. Next, the frustum is analysed for point measurements, if there are points in the frustum, the points are corroborated with ground truth labels to determine if they belong to a labelled object. We measure the accuracy (True Positives) of the shadow identification and object verification procedure by counting the total matched objects. We define False Negatives as labelled objects that we have missed, and False Positives as detected obstacles that do not match the labelled objects.

\vspace{3pt}\vspace{3pt}\noindent\textit{RQ2: How well does the methodology identify regions in a scene where an object is subjected to Object Hiding Attack?} We simulate Object Hiding Attacks and evaluate how well we can localize the regions in a scene where objects exists but are undetected by 3D Object Detectors. We use the procedure to identify shadow regions and clusters of shadows. For each shadow cluster, their discretized shadow space is sampled and the frustum from the LiDAR is generated to analyse for points in these frustums. If the points in the frustums do not belong to any detected objects, the un-labelled points are recorded. Then, all the un-labelled points are clustered to form clusters of un-labelled objects. We generate bounding boxes that enclose these clusters of points and effectively use them to localize the object. We then match the un-labelled objects to adversarially removed objects (from the KITTI ground truth) and calculate the Birds-Eye-View 2D Intersection-over-Union (IoU) of the ground truth object bounding boxes with those generated by localizing objects from their shadows. We measure the average IoU of all the hidden objects in the front-near region of interest from the ego-vehicle to evaluate the accuracy of the localization approach. Additionally, we measure the maximum distance of the nearest edge (to the ego-vehicle) of the bounding boxes.

\subsection{Shadow Identification and Object Attribution}
We run the proposed obstacle detection on 7480 scenes of the KITTI dataset. For every shadow detected, the frustums are generated and points in frustums are corroborated with ground truth bounding boxes of objects by checking if these points fall in the labelled bounding boxes. In the 7480 scenes, there are 10426 objects that are in the front-near region of interest. Out of these, 10259 (TPR 98.4\%)  were matched, showing a high accuracy in detecting objects. We analyzed the missed objects (FNR 1.6\%), and we found that the objects missed by the obstacle detection are due to these objects being located too close to the periphery of the RoI such the object is in the ROI but their shadows fall outside the analysis region. The current width of the RoI is chosen to be about 1.5$\times$ the width of a vehicle lane from the center of the ego-vehicle and hence the objects missed are at least 1 vehicle lane away, as the objects are closer to the center of the RoI, the shadows will be picked up and the obstacle would be detected. The width of the RoI is a parameter that can be increased if more visibility is required. We also recorded a False Positive Rate of 11.9\%, these are obstacles that are detected by the proposed approach but do not match the ground-truth labelled objects. Upon further analysis of these cases, we attribute the FPs to objects that are found commonly in the environment such as road dividers and traffic lights, as demonstrated in Section \ref{subsec:ex1}. Although the current approach is unable to identify obstacles and provide object labels, it is able to faithfully detect obstacles in a driving scene. 

\subsection{Localizing Object Hiding Attack Detection}
\begin{figure}[htb]
    \centering
    \subfigure[BEV of Bounding Boxes (Predicted and Ground Truth)]
    {
        \includegraphics[width=0.45\textwidth]{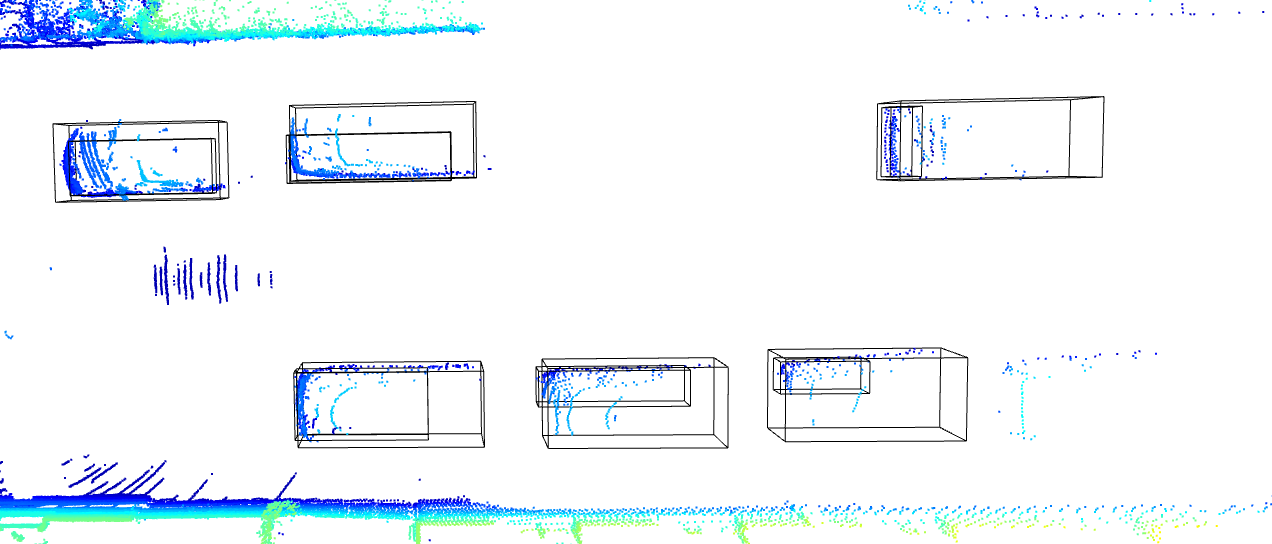}
        \label{fig:scene_2_iou1}
    }
    \subfigure[Another view of Bounding Boxes (Predicted and Ground Truth)]
    {
        \includegraphics[width=0.45\textwidth]{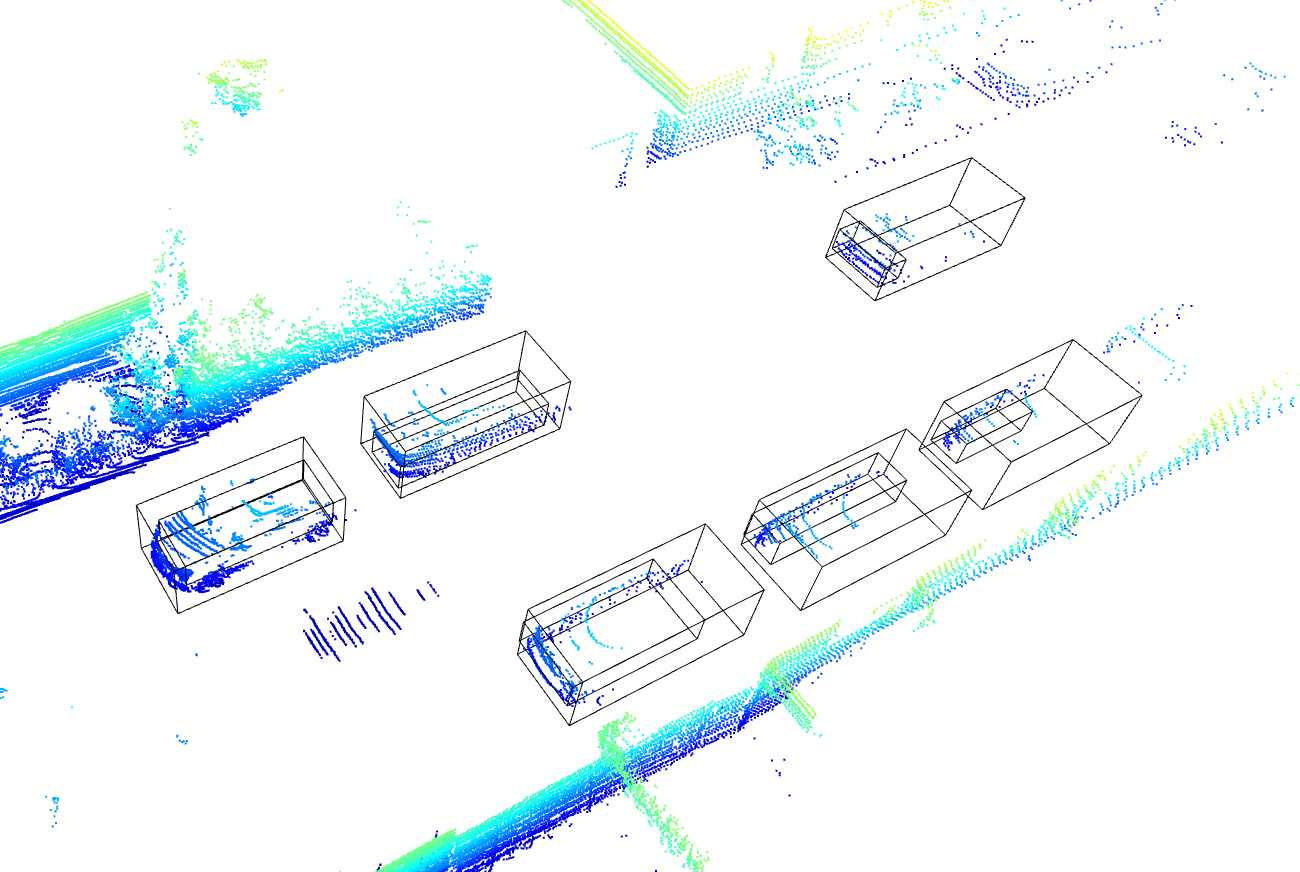}
        \label{fig:scene_2_iou2}
    }
    \caption{IoU example with a scene with multiple objects.}
    \label{fig:scene_2_iou}

\end{figure}

Next, we investigate how well the proposed obstacle detection is able to localize objects and compare the localization to ground truth labelled bounding boxes. In the 7480 scenes, obstacles were localized using the proposed obstacle detection and we measure the 2D BEV IoU of the predicted bounding box with the ground truth bounding boxes of objects in the front-near region. The average IoU for the bounding boxes of obstacles detected in the scenes is 33.2\%. The low average IoU can be attributed to the predicted bounding boxes only enclosing the point clusters found in the frustums (Fig. \ref{fig:scene_2_iou}). The smaller predicted bounding boxes are due to "missing points" that can arise from the DBSCAN clustering that label points that supposedly belonging to the object but are too far away as noise, and hence are not included into the clusters.

However, we observed that low IoU does not indicate that the predicted bounding boxes are not well localized, rather they are just smaller in size. As such, we used another metric where we measure the distance between the nearest edges (to the ego-vehicle) of the predicted and ground truth bounding boxes. This provides a metric of how well the obstacle detection predicts the distance of the obstacle from the ego-vehicle. We found that the average distance between the predicted and ground truth bounding box edges to be 1.8m (standard deviation = 2.4m), which shows that the prediction of the obstacle is very close to ground truth labels in terms of distance from the ego-vehicle.

\subsection{Runtime Efficiency}
We measure the runtime of the proposed obstacle detection using a machine equipped with an Intel Core i7 Six Core Processor i7-7800X (3.5GHz) and 32GB RAM. The average runtime for the full obstacle detection on 7480 scenes is 36.5 s/scene with a standard deviation of 7.7 s/scene. The majority of the processing time for the obstacle detection pipeline is spent on the two DBSCAN operations, first for clustering empty cells and second for clustering points in the frustums. As such, the current implementation is not ideal for use in real-time obstacle detection for autonomous driving decisions.

\subsection{Summary}
We have evaluated the utility and efficiency of the proposed obstacle detection methodology. We were able to detect 3D shadows and match them to the obstacles that caused them with a high accuracy (98.4\%). Although the proposed obstacle detection is unable to predict the correct size of the object, it can be used to estimate the obstacle's distance away from the ego-vehicle. Our current prototype implementation of our proposed obstacle detection mechanism was found to be not ideal for real-time processing of AV scenes and we are planning to work on optimisations to improving its efficiency.

\section{Discussion}
\label{sec:discussion}

\vspace{3pt}\vspace{3pt}\noindent\textbf{Detection approach.} Our proposed obstacle detection approach uses shadows as physical invariant of obstacles in a driving scene to locate these obstacles in a scene. An approach used by Autoware \cite{kato2018autoware} is to perform euclidean clustering on the points in a down-sampled point-cloud to detect objects, where its object detection result can be verified against the 3D object detector's output and a rule-based criteria can be used to determine if an object is being hidden. While augmenting Autoware's approach for detection is straight-forward, our approach of using shadows as an orthogonal defense is still useful as it provides a new perspective (looking at a physical invariant) and opportunities for better interpreting anomalies.  

\vspace{3pt}\vspace{3pt}\noindent\textbf{Inability to identify type of object.} Compared to DNN-based 3D object detection models, our proposed obstacle detection is unable to recognize the type of the object (i.e. unable to label the object). This is due to the use of clustering algorithms to cluster points that are in close proximity to each other to form clusters that represent the objects. The use of clustering approach does not have any notion of how various object types are represented in point cloud and hence, unable to predict the type of object. The clustering approach that group points by proximity also leads to the shortcoming of poor prediction of an object's size. Future work could look into how different modalities or machine learning approaches can be combined to enrich the information from shadows and detected obstacles for improving the semantics of our predictions.

\vspace{3pt}\vspace{3pt}\noindent\textbf{Real-time processing requirements.} Processing driving scenes in real-time is a critical requirement for autonomous driving decision making. Due to the computational overhead of our proposed approach, it is not suitable for real-time obstacle detection. However, there is still merit for the current approach to be used as an offline analysis tool that can aid in the detection of anomalous objects. Future work can be done to improve on the efficiency of the approach by exploring other clustering algorithms and localization strategies to optimise the performance. Another possible approach would be to have a lightweight implementation that identify an obstacle's location by only using the shadow clusters, doing away with the need of clustering points in frustums to generate the bounding boxes.

\vspace{3pt}\vspace{3pt}\noindent\textbf{Shadow removal adversary.} The threat model we consider in this work assumes that the adversary only performs object hiding attacks and does not take steps to conceal the attack against our proposed obstacle detection. An adaptive adversary can try to evade obstacle detection by performing LiDAR spoofing attacks to inject LiDAR point measurements into the void region of "hidden objects" to perturb their shadows. In future work, we plan to investigate the robustness of using shadows to detect obstacles against a stronger adversary that can simultaneously hide objects and perturb their shadows.

\section{Conclusion \& Future Work}
\label{sec:conclusion}
In this paper, we leveraged the key observation that objects from object hiding attacks leave behind shadow artifacts in their LiDAR 3D point cloud and this can be exploited to detect objects hidden from 3D object detection. To this end, we propose an obstacle detection methodology that first searches the front-near region of the ego-vehicle for void regions in the 3D point cloud and localizing obstacles that causes these shadows. We evaluated the performance of the obstacle detection methodology and found that it can match shadows to object with an accuracy of 98.4\%. It is also able to predict the distance of the obstacles to the ego-vehicle with an average precision of 1.8m. In future work, we hope to improve further on the utility of obstacle detection, looking into ways to increase the precision of the localization of obstacles and improving the efficiency to meet real-time processing requirements. We also plan to evaluate the robustness of using shadows for obstacle detection against an adaptive attacker that aims to remove shadows to evade detection.

\section*{Acknowledgment}
We would like to thank Qi Alfred Chen and the anonymous reviewers for their valuable feedback on our work.

\bibliographystyle{IEEEtran}
\bibliography{biblio}




\end{document}